\documentclass[letterpaper, 10 pt, conference]{ieeeconf}  

\IEEEoverridecommandlockouts                              

\usepackage{graphics} 
\usepackage{amsmath} 
\usepackage{amssymb}  
\usepackage{tikz}
\usepackage{stfloats}
\usetikzlibrary{arrows.meta,
                calc, chains,
                quotes,
                positioning,
                shapes.geometric}
\usepackage[x11names]{xcolor}
\usepackage{textcomp}
\usepackage{pgfplots}
\pgfplotsset{compat=1.6}
\usepgfplotslibrary{groupplots}
\usetikzlibrary{external}
\usepackage{fontawesome5}
\usepackage[caption=false,font=footnotesize]{subfig}
\usepackage{cases}
\usepackage{booktabs}
\usepackage{bm}
\usepackage{url}
\let\labelindent\relax
\usepackage{enumitem}
\usepackage{cleveref}
\usepackage{mathtools} 

\title{\LARGE \bf
Safety filtering of robotic manipulation under environment uncertainty: a computational approach
}

\author{Anna Johansson$^{1}$, Daniel Lindmark$^{2}$, Viktor Wiberg$^{2}$,  and Martin Servin$^{1,2}$
\thanks{*This work was supported in part by LKAB (AC-438), Mistra Digital Forest, Sweden Grant DIA 2017/14 \#6, and Horizon Europe Project XSCAVE under Grant 101189836.}
\thanks{$^{1}$Department of Physics, Umeå University, Umeå, Sweden.}%
\thanks{$^{2}$Algoryx Simulation AB, Umeå, Sweden.}%
}

\begin{document}

\maketitle
\thispagestyle{empty}
\pagestyle{empty}

\begin{abstract}
        Robotic manipulation in dynamic and unstructured environments requires safety mechanisms that exploit what is known and what is uncertain about the world. Existing safety filters often assume full observability, limiting their applicability in real-world tasks. We propose a physics-based safety filtering scheme that leverages high-fidelity simulation to assess control policies under uncertainty in world parameters. The method combines dense rollout with nominal parameters and parallelizable sparse re-evaluation at critical state-transitions, quantified through generalized factors of safety for stable grasping and actuator limits, and targeted uncertainty reduction through probing actions. We demonstrate the approach in a simulated bimanual manipulation task with uncertain object mass and friction, showing that unsafe trajectories can be identified and filtered efficiently. Our results highlight physics-based sparse safety evaluation as a scalable strategy for safe robotic manipulation under uncertainty.
\end{abstract}

\section{Introduction}

The growing deployment of autonomous robots beyond traditional assembly-line settings presents an increasing need for control schemes capable of handling complex and dynamic environments while guaranteeing both task success and safety. Safety filters address this need by monitoring the agent's proposed actions and, when necessary, restricting them to a safe action space~\cite{annurev:/content/journals/10.1146/annurev-control-071723-102940}.

However, in many settings where autonomous robots are particularly useful, such as heavy lifting and object manipulation in remote or hazardous environments, the physical properties of the environment are often partially unknown or uncertain \cite{Wong01012018}. This presents unique challenges for safe motion planning, as most existing safety filters assume perfect world observability \cite{annurev:/content/journals/10.1146/annurev-control-071723-102940}. While prior work has explored safe control under environmental uncertainty, there is a lack of work addressing the unique challenges that uncertainty poses in object manipulation tasks. These challenges include accounting for the inherently complex contacting multi-body dynamics of manipulation tasks, as well as the necessity of interacting with unknown materials and objects during task execution. Consequently, object manipulation is fundamentally different from many other established safe control domains, such as autonomous driving. To address this gap, we investigate the feasibility of leveraging the predictive capabilities of a physics engine to develop a physics-based safety monitoring method which can be used for ensuring safe rollout of a control policy at a given world state. Our approach is to evaluate the effects of the corresponding action sequence under environmental uncertainty using high-fidelity physics simulations, thus providing a basis for evaluating the safety and for improving it through better world knowledge.

However, simulating rollouts for each possible world configuration is computationally intractable. To address this, we reduce the computational load by solving the system dynamics only at a few critical points in time for most of the tentative world parameter values.

This is accomplished by solving the system's dynamics in two stages. First, we compute a nominal state-action sequence through a simulated rollout with some nominal values for all uncertain world parameters. Second, we select critical states identified from the rollout and re-evaluate them with alternative world parameters. In this stage, the system dynamics is solved only once (a single state transition)  for each combination of tentative parameter values.

This strategy builds an understanding of the possible outcomes of the control policy without requiring excessive computational resources, and since all the calculations are independent, we can run them in parallel to increase speed. By simulating the consequences of the proposed action across a wide range of plausible world configurations, the method enables a robust safety assessment and filters out unsuitable actions.

In this paper, we demonstrate and evaluate this approach in a simulated scenario. In addition, we perform a computational feasibility analysis, to explore the viability of using it in real life where the uncertainty of the environment might be larger than in our limited example. With this, we examine whether the approach can be transferred to a real robot performing manipulation tasks in uncertain environments, with the goal of ensuring safe control.

\section{Background}
This section reviews related work on safety filtering and the computational framework on which our method is built.

\subsection{Safety Filters}
Safety filters are a family of methods which aim to ensure that autonomous control results in only safe behavior by monitoring and, if needed, modifying the planned actions at runtime \cite{annurev:/content/journals/10.1146/annurev-control-071723-102940}. A common approach to safety filtering is to define sets of states as either failure sets or safe sets. A failure state is often defined in such a way as to include all states from which catastrophic failure is inevitable. The goal of the filter then, is to make sure that the system does not enter such a failure state. 

One way to construct the failure sets is through Hamilton--Jacobi reachability analysis, which enables the computation of all states which will lead to future unsafe states, and thus enables the avoidance of such failures. One problem with this approach is that the calculations become intractable at higher state dimensions. Approaches to combat this include leveraging reinforcement learning to approximate solutions \cite{7039601, 8794107, nakamura2025generalizingsafetycollisionavoidancelatentspace}. Hamilton--Jacobi--Isaacs analysis formulates the problem as a zero-sum-game where one safety oriented agent and one adversarial agent are competing. This allows one to take into account the worst-case disturbances and uncertainties with the safety filter \cite{hsunguyen2023isaacs, nguyen2025gameplayfiltersrobustzeroshot}. 

Other approaches to safety filtering includes using control barrier functions \cite{CBF} and model predictive control \cite{WABERSICH2021109597}. In \cite{WABERSICH2021109597}, they construct a safety filter based on a probabilistic model of the system, which is inferred from data. The filter is constructed so that it can be used online without the need of the heavy offline calculations necessary when using for example Hamilton--Jacobi reachability analysis.

While many safety filters focus on collision avoidance, there are some works on avoiding other kinds of failures. \cite{nakamura2025generalizingsafetycollisionavoidancelatentspace} uses RL to learn latent safety constraints to avoid failures which are hard to formulate analytically, like spilling the contents of a bag. A data-driven MPC-based safety filter is proposed in \cite{9555680}. It uses Gaussian process models to approximate safe sets of a robot pushing a box. Specifically, scenarios in which the box slides are labeled as safe, whereas cases in which the box remains fix are classified as unsafe. In \cite{seo2025uncertaintyawarelatentsafetyfilters}, a world model is trained on robot-environment interaction data. The data is labeled with failure labels and the safety filters are computed using Hamilton--Jacobi reachability analysis on these latent safety constraints. Within this work, unsafe sets corresponds to situations in which the robot’s actions either are known to  result in a Jenga tower collapsing or are out-of-distribution. In \cite{TANG20237912}, reachability analysis was utilized to verify that a planned trajectory was safe with respect to impacts, while accounting for time- and input uncertainty.

Avoiding slip in manipulation task have so far largely been treated as a separate issue from safety filtering, but some studies have sought to mitigate slip by training models to anticipate slip and act preventative when need arises \cite{7354090, Nazari2025}, an approach which shares some similarities with safety filtering. 



\subsection{Simulation}\label{sec:simulation}
In robotics, simulations are usually based on discrete-time multibody dynamics with numerical algorithms for time stepping and equation solving implemented in a physics engine.
In descriptor form (full-coordinate formulation), the state of the system is represented by the positions $\bm x_k$ and velocities $\bm v_k$ of the involved bodies at a discrete time $k$.
Joints, driveline, actuators, and contacts are represented by kinematic constraints, $\phi(\bm x_k,\bm v_k)=0$, which give rise to constraint forces with magnitude given by their Lagrange multipliers $\bm \lambda_k$.
Actuator and driveline settings can be controlled via input signals $u_k$.
The time evolution of the dynamic system is the result of its initial state, the involved constraints, external forces $f_k$ (such as gravity), and the control signal.
A simulation is a rollout of the multibody dynamics given some control policy $\pi$ 
\begin{align}
    \bm z_{k+1} & = f(\bm z_k,\bm u_k; \bm \theta) \\
    \bm u_k & = \pi(\bm z_k)    
\end{align}
where $\bm z_k = [\bm x_k, \bm v_k, \bm \lambda_k]$ is the state of the system augmented with the Lagrange multipliers and $\bm \theta$ is the world parameters. The latter may include inertial parameters, link dimensions, gravitational acceleration, friction coefficients, motor gains, deadbands etc. 
The function $f$ represents the time-stepper and equation solver.
With semi-implicit time-stepping and constraint linearization, one obtains the following generic integrations scheme, manifested in many physics engines
\begin{equation}
    \label{eq:x-step}
    \bm x_{k+1} = \bm x_k + \bm v_{k+1} \Delta t
\end{equation}
\begin{equation}
    \label{eq:v-step}
    \begin{bmatrix}
        \bm M_k & -\bm G_k \\
        \bm G_k & \bm \Sigma 
    \end{bmatrix}
    \overbrace{\begin{bmatrix}
         \bm v_{k+1}\\
         \bm \lambda_{k+1}
    \end{bmatrix}}^{ \bm y}
    - \begin{bmatrix}
         \bm r_v\\
         \bm r_\lambda
    \end{bmatrix}
    =  \bm w_l -  \bm w_u
\end{equation}
\begin{equation}\label{eq:cc}
    \begin{gathered} 
        0 \leq \bm y - \bm l \perp \bm w_l \geq 0 \\
        0 \leq \bm u - \bm y \perp \bm w_u \geq 0,
        \end{gathered}
\end{equation}
where $\bm M$, $\bm G$, are block-sparse matrices with inertial parameters and constraint Jacobians, and $\bm \Sigma$ is a diagonal regularization matrix, respectively.
In Eq.~\eqref{eq:v-step},  $\bm r_v =  \bm M  \bm v_k + \Delta t  \bm M^{-1} \bm f_k$ and $ \bm r_\lambda(\bm x_k,\bm v_k,\bm u_k)$ depend on the specific method for constraint stabilization. Eqs. \eqref{eq:v-step}-\eqref{eq:cc} form a linear complementarity problem (LCP) \cite{murty1988linear} with slack variables $\bm w_l$ and $\bm w_u$, and upper and lower limits $\bm u$ and $\bm l$ that depend on system parameters such as joint limits, actuator limits, and friction coefficients. The complementarity conditions from frictional contacts involve unilaterality of the normal force multiplier, 
$\lambda_\text{n} \leq 0$, and the friction force multiplier obeying the Coulomb law, $\lambda_\text{t} \leq \mu \lambda_\text{n}$, enforcing zero slip when below the bound set by the friction coefficient $\mu$.

We use the AGX Dynamics \cite{AGX} physics engine with support for the SPOOK stepper \cite{SPOOK}, optimized block-pivot direct solver for high precision for ill-conditioned systems, and iterative projected Gauss-Seidel solver for approximate solutions on large contact problems. 
We prefer this choice over, for example, a physics engine based on position-based dynamics or reduced coordinates since safety filters require high observability and precision in constraint forces at articulation joints, motors, and contacts, and numerical capability to handle kinematic loops and large ratios in masses and stiffness. AGX Dynamics has proven capable of simulating and providing sim-to-real transfer for complex machinery in dynamic environments \cite{wiberg2024,aoshima2024}.

\section{Method}
We explore a safety monitoring scheme for tasks with parameter uncertainty by monitoring the expected effect of rolling out a control policy, $\pi$, in a high-fidelity simulator. Our rationale for this approach is that physics-based simulation allows for highly realistic safety evaluations of complex dynamics, assuming the simulator has been pre-calibrated, except for the uncertain world parameters specific to each scenario. 
Evaluating the entire proposed action sequence across the full range of possible world parameter values would be computationally prohibitive. To address this, we introduce a method of \textit{sparse evaluation}, in which the safety of the planned sequence is assessed only at identified critical state-transitions, where the risk of failure appears to be the greatest. We argue that accounting for these critical state-transitions is necessary for deciding if the entire trajectory is safe or not, in a similar vein as the Hamilton--Jacobi--Isaacs analysis \cite{hsunguyen2023isaacs, nguyen2025gameplayfiltersrobustzeroshot}. This approach yields a probabilistic bound on safety violations, allowing specification of an acceptable risk level while ensuring computational feasibility through minimized calculations.

The complete pipeline of the method is detailed in this section, with each subsection describing different components. A graphical illustration of the proposed pipeline is provided in Fig. \ref{fig:flow_chart}. The formulation of a safe fall-back policy $\pi^{\text{\tiny \faShield*}}$ is outside the scope of this paper. Our contribution is a method for anticipating potential failures under environmental uncertainty, as a first step towards preventing it.

\usetikzlibrary{fit}  
\begin{figure}[t]
\vspace{2mm}
    \begin{tikzpicture}[
        node distance = 10mm and 13mm, 
        font=\small,
        start chain = A going below,
        base/.style = {draw, minimum width=15mm, minimum height=6mm,
                        text width=25mm, align=center, on chain=A},
        start/.style = {base, rectangle, rounded corners, fill=blue!15},
        red/.style = {base, rectangle, rounded corners, fill=red!30},
        green/.style = {base, rectangle, rounded corners, fill=Green2!30},
        process/.style = {base, rectangle, fill=gray!25},
        decision/.style = {base, diamond, aspect=1, text width=12mm, fill=Gold1!30},
        comment/.style = {base, text width=15mm, rounded corners, dashed},
        every edge quotes/.style = {auto=right, font=\scriptsize}]
                        ]
        \node[start] (1) at (0,0) {Nominal control policy $\bm u_k =   \pi(\bm z_k)$};
        \node[start, left=3mm of 1] (B) {Known initial state $\bm{z}_0$ and system dynamics $f$};
        
        \node[fit=(1)(B), inner sep=0, draw=none] (group) {};
        
        \node[process, below=of group] (2) {Simulator};
        
        \node[fit=(1)(B), inner sep=0, draw=none] (group) {};
        \node[base] (3) {Nominal state-action sequences $\bar{\bm{z}}_{0:H}$, $\bm{u}_{0:H}$};
        \node[base] (5) {$ \mathcal{C} = \{c_1, \dots, c_K\}$};
        \node[process, text width=26mm] (6) {Solver\\{\scriptsize $\bm{z}_{c+1}^*=f(\bar{\bm{z}}_c, \bm u_c; \bm \theta^*)$}};
        \node[process] (7) {Safety Evaluation};
        \node[decision] (8) {Acceptable safety score?};
        \node[decision, right=of 8] (9) {Large\\uncertainty $p(\bm\theta)$?};
        \node[process] (10) at ($(9 |- 7) + (0,11mm)$) {Safe probing action};
        \node[decision, above=of 10, yshift=-3mm] (11) {$\bar{\bm\theta}' \approx \bar{\bm\theta}$?};

        %
        \node [base, right=of 2, xshift=10pt, text width=15mm] (nominal params) {Updated $\bar{\bm \theta}'$, $p'(\bm\theta)$.};
        \node [comment, left=of 2, xshift=35pt,text width=15mm] (tent params) {Tentative $\bar {\bm{\theta}}$, $p(\bm{\theta})$.};

        
        \node[green, below=of 8] (safe) {Roll out nominal policy};
        \node[red, below=of 9] (unsafe) {Switch to\\safe policy $\pi^{\text{\tiny \faShield*}}$};
        \draw [arrows=-Stealth] 
          (1) edge (2)
          (B) edge (2)
          (2) edge["Dense rollout"'] (3)
          (3) edge["Identify critical states"'] (5)
        
          
          (5) edge[transform canvas={xshift=-18pt}, dotted] (6)
          (5) edge["\dots"', transform canvas={xshift=-12pt}, dotted] (6)
          (5) edge[transform canvas={xshift=5pt}, dotted] node[xshift=15pt]{} (6)

          (6) edge[transform canvas={xshift=-18pt}, dotted] (7)
          (6) edge["\dots"', transform canvas={xshift=-12pt}, dotted] (7)
          (6) edge[transform canvas={xshift=5pt}, dotted, "$\bm{z}^*_{c+1}$"'] (7)

          (7) edge (8)
          
          (8) edge["No"] (9)
          (10) edge node[midway, right, font=\scriptsize, align=left] {$p'(\bm\theta)$} (11)
          (9) edge["No"] (unsafe)
          (9) edge["Yes"] (10)

          (11) edge node[midway, right, font=\scriptsize, align=left] {No:\\Redo dense\\evaluation} (nominal params)
          (11) edge node[midway, below, font=\scriptsize, align=center, xshift=5pt] {Yes: \\ reevaluate\\at same $c$, \\ for updated $p'(\bm\theta)$ } (6)

          (tent params) edge (2)
          (tent params) |- (6)

          (nominal params) edge (2)
          (8) edge["Yes"] (safe)
          ;
    \end{tikzpicture}
    \caption{A flowchart illustrating our proposed pipeline for a safety filtering method.}
    \label{fig:flow_chart}
\end{figure}
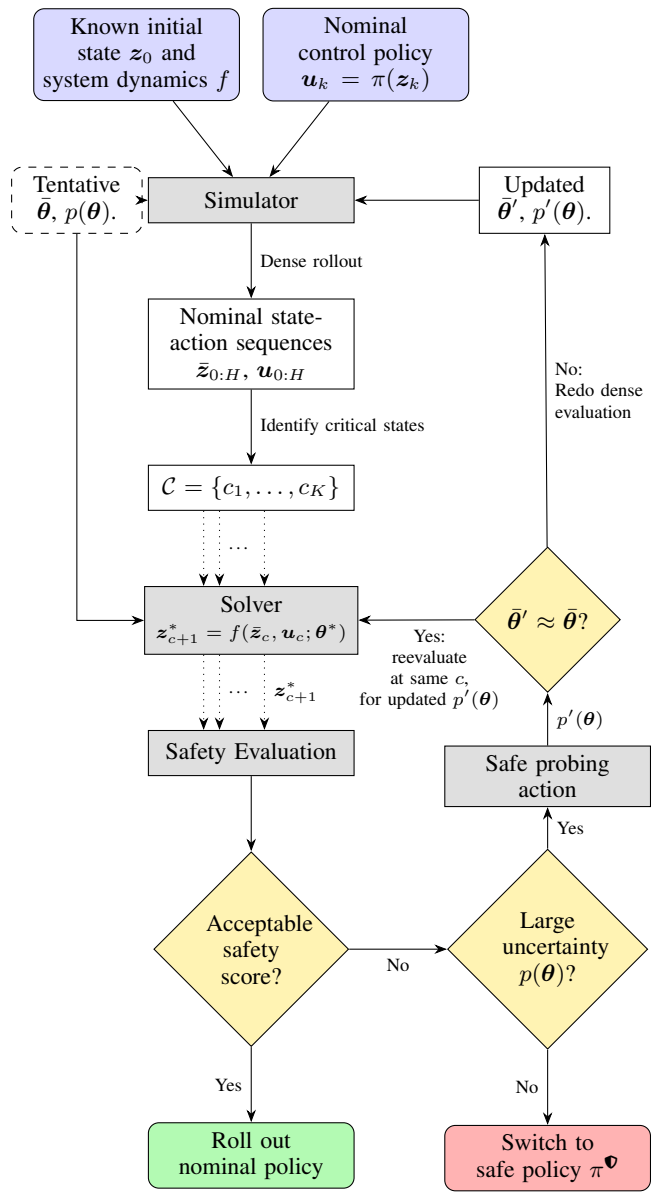

\subsection{World Knowledge}
To enable predictions of the states resulting from the planned action sequence, the world is represented through a simulator, $\bm z_{k+1} = f(\bm z_k,\bm u_k; \bm \theta)$, as described in Sec.~\ref{sec:simulation}. The simulator contains information about all bodies, actuators, and the dynamics of the robot and its surrounding environment. 
We assume perfect knowledge of the robot, but that the environment is only partially known. Specifically, we assume that there is a set of $N$ environmental parameters, $\bm \theta \in \mathcal{D} \subseteq \mathbb{R}^N$ with significant uncertainty,
represented by a probability distribution $p(\bm \theta)$, within some bounds $\mathcal{D}$. From this, one can extract nominal parameters, for example $\bar{\bm \theta} = \mathbb{E}\left[p(\bm \theta)\right]$, and then run a simulation. 
We assume that the initial state $\bm{z}_0$ is known and that a (deterministic) control policy $\bm u_k = \pi(\bm z_k)$ is given.

\subsection{Dense Evaluation using Nominal Parameters} \label{sec:Dense_eval}
As the first step in our proposed evaluation method, a roll-out of the control policy is carried out in the simulated environment. In this first rollout, which we call the \textit{dense evaluation}, nominal values are used for the uncertain parameters of the system. This results in a series of nominal states, a trajectory $\bar{\bm{z}}_{0:H}$  of length $H$, where $\bar{\bm{z}}_0 = \bm{z}_0$ and
\begin{equation}
    \bar{\bm{z}}_{k+1} = f(\bar{\bm{z}}_k, \bm u_k; \bar {\bm \theta}).
\end{equation}
The safety margin of each state-transition during this rollout is then examined. To this end, we employ the concept of Factor of Safety (FOS) to quantify the system’s margin to failure. In mechanical engineering, FOS is the ratio of the maximum strength of a component over the expected maximum load during operation. We generalize this notion to quantify whether a grasp is close to slipping, depending on the frictional contact forces, or whether an actuator is close to reaching its torque limit. 
A value of $\mathrm{FOS} = 1$ corresponds to a definite risk of failure, whereas $\mathrm{FOS} \gg 1$ indicates a substantial safety margin. For normalization purposes, we use the inverted FOS in this paper, so that $\mathrm{FOS^{-1}} \in [0,1]$, where a low score indicates safety. The exact formulation depends on the specific task and the risks it poses, but in general, the factor of safety can be expressed as a function of the current state:
\begin{equation}
    \mathrm{FOS}^{-1}_k = g(\bm{z}_k).
    \label{eq:FOS}
\end{equation}
We evaluate the factor of safety for every nominal state $\bar{\bm{z}}_k$, and, as an initial safety check, examine the scores to decide whether or not the nominal trajectory is safe and if the task is successful under nominal parameter values. If safety cannot be ensured under the nominal parameter values, we invoke a safe fallback policy $\pi^{\text{\tiny \faShield*}}$. Otherwise, the evaluation proceeds to the next stage of the pipeline. 
\subsection{Sparse Evaluation}
In this stage, we evaluate the safety of the policy rollout under parameter uncertainty. Dense evaluation for each tentative $\bm \theta \in \mathcal{D}$ is computationally expensive, making an exhaustive search intractable. Instead, we reduce the number of necessary computations by identifying the state-transitions with the highest risk of $\bm{z}$ entering a failure set and evaluating safety only at those instances. We refer to this as a sparse evaluation tactic. These critical state-transitions are determined from the dense evaluation by examining the inverse safety factor, $\mathrm{FOS}^{-1}$, where larger values indicate higher risk of failure. 

To assess the critical state-transitions $c \in \mathcal{C} \subseteq \mathbb{N}$ under system uncertainty, we re-evaluate the dynamics with
\begin{equation}
    \bm{z}_{c+1}^* = f(\bar{\bm{z}}_c, \bm u_c; \bm \theta^*),
\end{equation}
where $\bar{\bm{z}}_c$ is a critical state, and $\bm \theta^*$ is sampled according to the parameter probability distribution $p(\bm \theta)$. By repeating this process over a sufficiently large set of samples, we obtain a representative characterization of the possible states that may evolve from $\bm{z}_{c}$. Since these calculations are independent of each other, they can easily be parallelized.

To evaluate the safety of each critical state-transition under parameter uncertainty, we compute a corresponding safety score $S_c$ for each such instance. The safety score is calculated as the integral over the uncertainty span of $\theta$ according to
\begin{equation}
    S_c = \int_{\bm \theta\in\mathcal{D}} p(\bm \theta) g(\bm{z}_c) \mathrm{d}\bm \theta.
\label{eq:safety_eval}
\end{equation}
In this way, the safety score reflects both the underlying uncertainty and the FOS, allowing situations with high uncertainty to still be considered safe if the factor of safety indicates sufficient confidence. The safety score for the critical state-transition is then evaluated by checking if the following statement holds true:
\begin{equation}
    S_c < \epsilon,
\label{eq:safety_criteria}
\end{equation}
for some acceptable safety tolerance $\epsilon$. If this criteria is met for every $c$, the proposed action is deemed safe under the environment uncertainty, and thus it can be rolled out on the real robot. If the criteria is not met for some $c$, we interpret this as an indication that the corresponding action may be unsafe in real-world execution. This leads to one of two possible next steps: uncertainty reduction or safe policy activation.

\subsection{Uncertainty Reduction}
When the uncertainty span is large for some parameter, we can reduce it by performing a safe probing action using the real robot. The outcome then updates the probability distribution to $p'(\bm \theta)$. If the nominal parameter value is close to the previous one, $\bar{\bm \theta} \approx \bar{\bm \theta}' = \mathbb{E}\left[p'(\bm \theta)\right]$, we assume that the previously identified critical state-transition remains representative. In this case, we simply re-evaluated the safety score using Eq.~\eqref{eq:safety_eval}. If the new nominal value differs significantly from the old one, we instead perform a dense evaluation with the updated nominal parameters, followed by a corresponding sparse evaluation. Depending on the updated safety score, the robot either executes the planned action in real life, or falls back to the safe policy $\pi^{\text{\tiny \faShield*}}$.


\section{Numerical Experiments}
In this section, we provide an example of how our method could be used together with an autonomous control policy. We describe the specific setup, safety criterias, and report results from the safety evaluations.

\subsection{Setup}
To test and analyze our method, we evaluate the safety when rolling out a policy from the imitation-learning based model ACT \cite{zhao2023learningfinegrainedbimanualmanipulation}. We use the same experimental setup as in the original paper to avoid additional training. For the safety evaluation, simulations are conducted with AGX Dynamics \cite{AGX}, which implements the multi-body dynamics formulations outlined in the Background section.

\begin{figure}
\vspace{2mm}
    \centering
    \begin{tikzpicture}
        \node[anchor=south west,inner sep=0, opacity=1] at (0,0)
            {\includegraphics[width=0.7\linewidth, trim=50 450 50 500, clip]{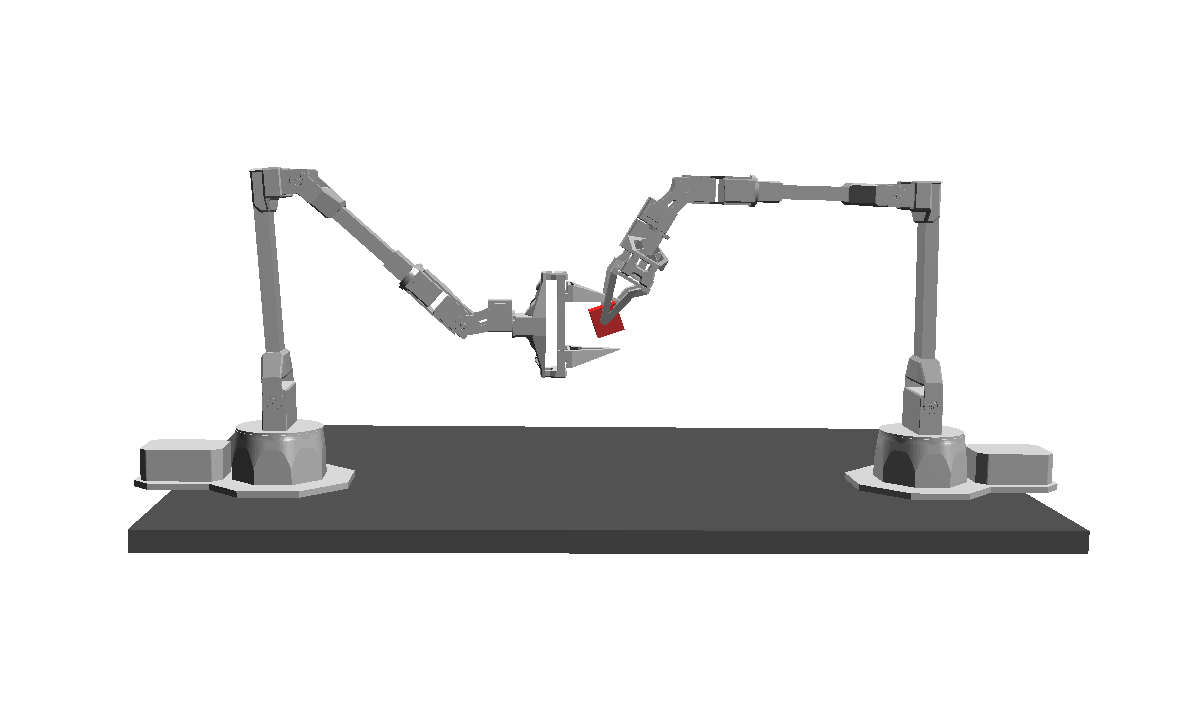}};
        \node[anchor=south west,inner sep=0, opacity=0.5] at (0,0)
            {\includegraphics[width=0.7\linewidth, trim=50 450 50 500, clip]{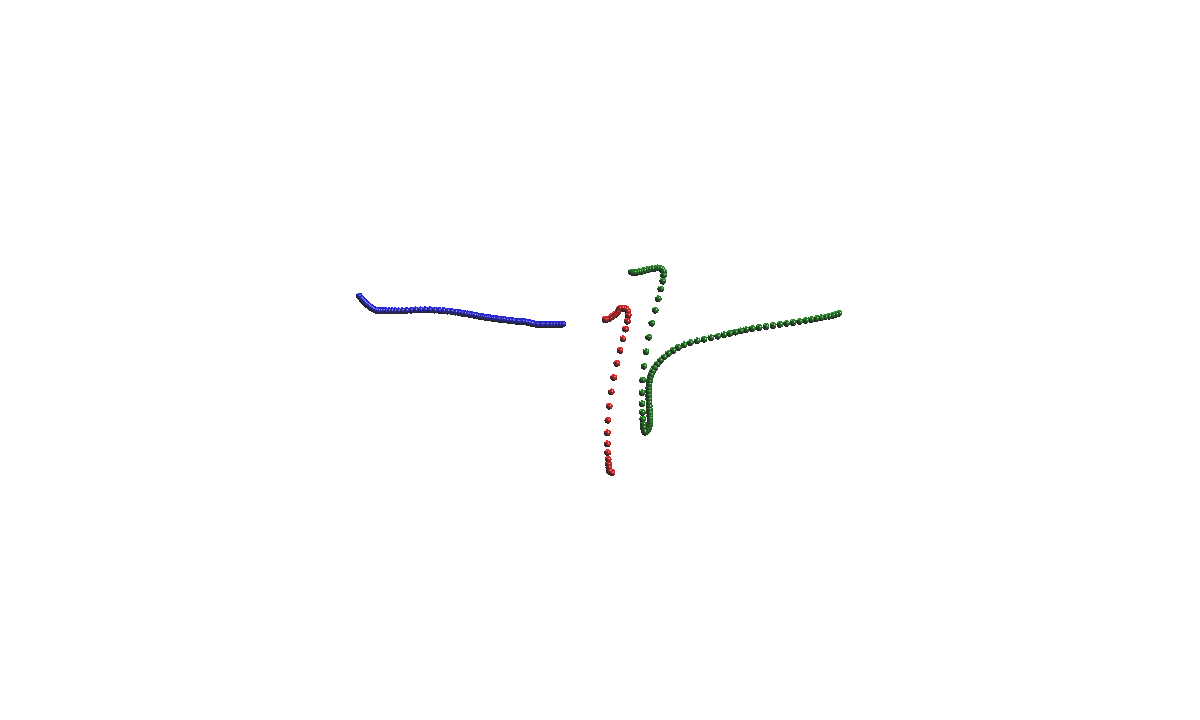}};
        \node[anchor=south west,inner sep=0, opacity=0.5] at (0,0)
            {\includegraphics[width=0.7\linewidth, trim=50 450 50 500, clip]{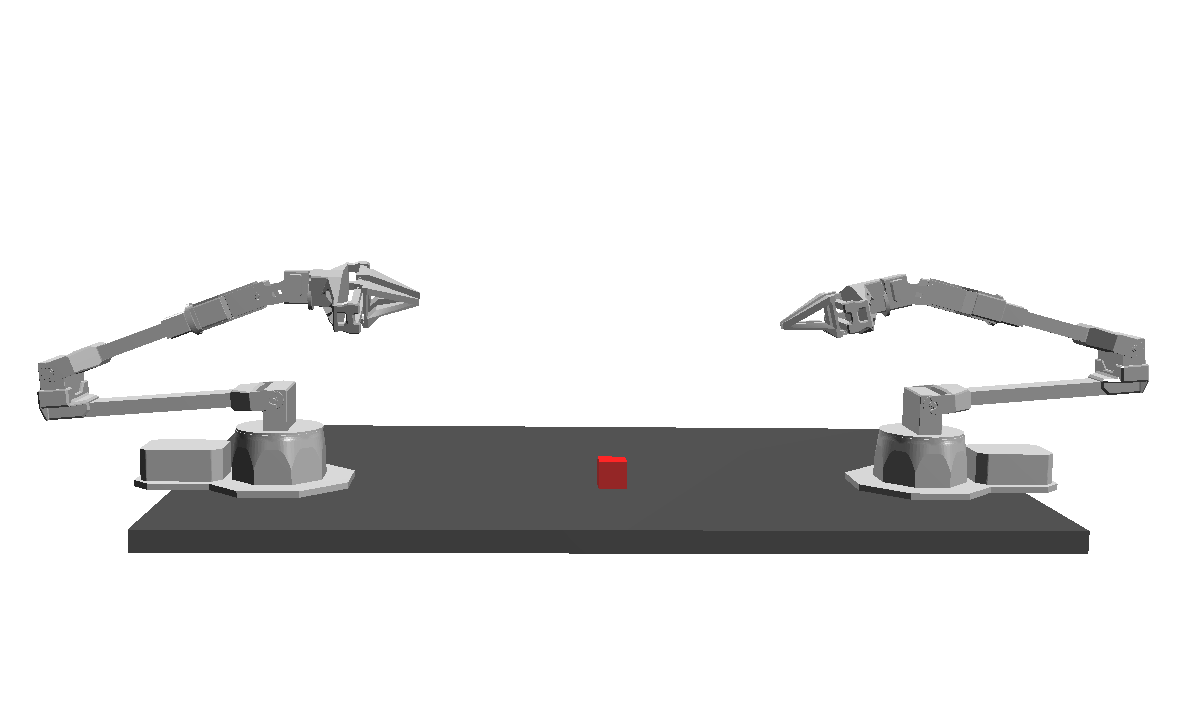}};
    \end{tikzpicture}
    \caption{A snapshot of the task at hand-over, where the path taken by the box and the two end-effectors are traced for visual representation of the trajectory. Transparent bodies depict initial poses.}
    \label{fig:setup_2}
\end{figure}

The manipulation task involves two ViperX 300 S robotic arms situated across from each other on a table, each equipped with parallel grippers. A box sits between the robotic arms, and the task is for one arm to pick it up and hand it over to the other. Fig. \ref{fig:setup_2} shows this setup as simulated using AGX Dynamics.

In our experiments, the unknown parameters are limited to the mass of the box and the friction coefficient between the box and the grippers, so that $\bm \theta = [\theta_m, \theta_\mu]$, with the entries corresponding to the mass and friction, respectively. For the nominal dense evaluation, we consider the nominal parameters $\bar{\bm \theta} = [0.25, 0.5]$ and assume a probability density function $p(\bm \theta)$ for these parameters given by a truncated Gaussian distribution normalized over $\bm \theta \in \mathcal{D}$. The density is centered at the nominal values, and the standard deviations are $\sigma_m = 0.8$ and $\sigma_\mu = 0.5$ for the mass and friction, respectively. The standard deviations are chosen to represent a wide range of uncertainty, to show how the method performs in such cases.

We evaluate combinations of these parameters on a grid with size $48 \times 48$.
\begin{figure}[t]
    \centering
    \subfloat[Critical state for $\mathrm{FOS}_{\mathrm{contact}}$\label{fig:setup_pick_up}]{%
       \includegraphics[width=0.49\linewidth]{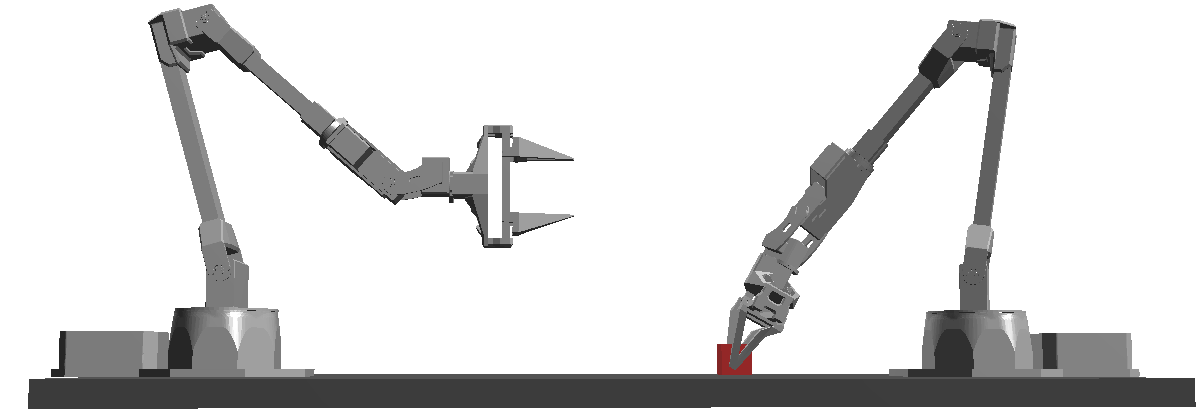}}
    \subfloat[Critical state for $\mathrm{FOS}_{\mathrm{motor}}$\label{fig:setup_hand_over}]{%
        \includegraphics[width=0.49\linewidth]{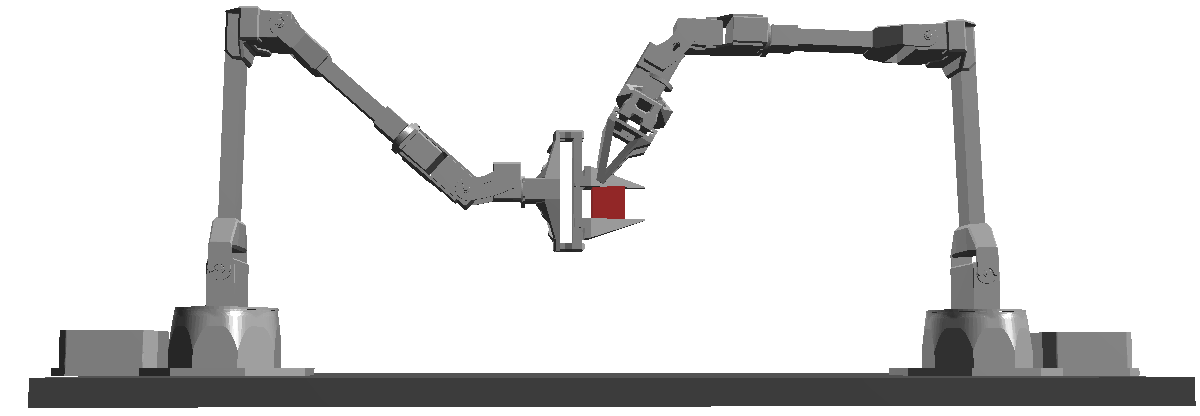}}
    \caption{Snapshots from the dense evaluation of the two critical states identified this paper, using the nominal world parameter distribution.}
    \label{fig:setup_critical_moments} 
\end{figure}
For this particular task, two safety risks are identified, namely: that the box can slip out of the grippers' grasp and that the actuators fail in executing the control action if the load becomes too heavy. With this background, two FOS functions are constructed. The first one is based on the Coulomb friction model. It is constructed so that the FOS approaches 1 when a specific contact enters slip-mode. The relevant contact for this task is the contact between the grippers and the box, which is represented by $M$ contact points in the simulator. This FOS is defined as
\begin{equation}
\mathrm{FOS}^{-1}_\mathrm{contact} =
\begin{cases}
    \scriptstyle
      \frac{1}{\sum^M_j \lambda_n^j} 
      \sum^M_i \lambda^i_n 
      \left( \frac{\lambda^i_t}{\mu^i \lambda^i_n} \right), &  r \notin \{1,3\}
     \\
    0, & \text{otherwise}.
\end{cases}
\end{equation}
Here, $\mu$ is the friction coefficient in the contact, and $\lambda_n$ and $\lambda_t$ is the magnitude of the normal force and the tangential force, respectively. Thus, the expression inside the parentheses can be interpreted as a local factor of safety against slip at contact point $i$. This expression is weighted with the normal force magnitudes so that contact points with small normal forces do not get an undue influence over the total FOS.

The variable $r$ tracks the accumulated reward of the trajectory, and is defined in such a way that $r=1$ when the right-hand robot has established contact with the box but not yet lifted it from the table, and $r=3$ when the left-hand robot has established contact with the box but the right-hand robot has not yet let it go. Defining $\text{FOS}^{-1}_\mathrm{contact}$ in this way allows for slip in the contacts during contact events, when it might be necessary and therefore should not be classified as unsafe. The precise definition of $r$ can be found in Appendix \ref{app:reward_def}.

To monitor the risk of actuator saturation, we formulate a FOS that indicates how close the robot is to reaching the torque limits of the joints currently engaged in the manipulation. A value of $\mathrm{FOS}^{-1}_{\mathrm{motor}} = 1$ means that at least one actuator operates at its maximum effort, indicating that the planned action sequence might be infeasible. We define it as
\begin{equation}
    \mathrm{FOS}^{-1}_{\mathrm{motor}} = \min_{j \in \mathcal{J}} \frac{|\lambda_j|}{\lambda_j^{\max}},
\end{equation}
where $\mathcal{J}$ is the set of actuators which are currently engaged in the manipulation, $\lambda_j$ is the torque applied by actuator $j$, and $\lambda_j^{\max}$ represents its maximum allowable torque. 

A combined measure on the total factor of safety is defined as
\begin{equation}
    \mathrm{FOS}^{-1} = \max \left[\text{FOS}^{-1}_\mathrm{contact}, \mathrm{FOS}^{-1}_{\mathrm{motor}} \right],
\end{equation}
such that the formulation accounts only for the most critical of the two contributions. This ensures that the system is classified as unsafe whenever a single risk component attains a sufficiently high level. This is the function we use as $g(\bm{z})$ when evaluating the safety score in Eq.~\eqref{eq:safety_eval}. Moreover, we set the safety tolerance in Eq.~\eqref{eq:safety_criteria} to $\epsilon = 0.75$.

Our experiments focus on demonstrating the proposed evaluation pipeline, since we do not currently have access to a suitable fallback policy. To emulate a probing action, we artificially update the parameter distribution and re-evaluate the newly identified critical state-transitions. Specifically, we use the same truncated Gaussian distribution as before, but with updated nominal parameters $\bar {\bm \theta}' = [0.2, 0.8]$ and reduced standard deviations $\sigma_m = 0.1$ and $\sigma_\mu = 0.2$. We refer to this updated function as $p'(\bm \theta)$.

For conciseness, we report only two critical state-transitions of the proposed trajectory in this paper. In a real-life application of the method for safety monitoring, it is likely necessary to examine additional state-transitions. 

\subsection{Results Dense Evaluation}
\begin{figure}[th]
\vspace{2mm}
    \centering
    \includegraphics[width=1\linewidth]{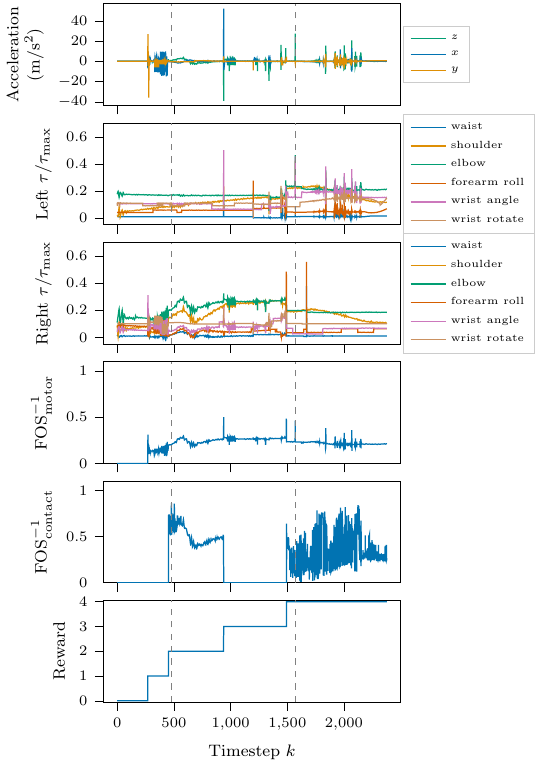}
    \caption{Results from the dense evaluation corresponding to the tentative nominal parameters probability distribution $p(\bm \theta)$. The two critical state-transitions on which the sparse evaluation is carried out are marked by gray dashed lines.}
    \label{fig:results_dense_eval}
\end{figure}

A dense evaluation using nominal parameter values $\bar{\bm{\theta}}$ is carried out, generating a series of nominal states. The two critical state-transitions are chosen by identifying nominal states at which at least one of the two FOS functions reaches critical values, and sampling one instance corresponding to each of these factors. Fig. \ref{fig:results_dense_eval} shows the monitored factors of safety, along with the acceleration of the box in world coordinate directions, as well as the fractions of the available torques $\tau/\tau_{\max}$ used by the respective actuators of the left and right robots, at every point in time in the dense evaluation. In these plots, the two critical states are indicated with gray vertical dashed lines. The reward, which is used to differentiate each phase of the task, is defined in Appendix \ref{app:reward_def}.

Fig. \ref{fig:setup_critical_moments} shows snapshots of the simulations at these two instances. The first, corresponding to $\mathrm{FOS}^{-1}_\mathrm{contact}$, occurs when the right-hand robot lifts the box toward the transfer point. The second, corresponding to $\mathrm{FOS}^{-1}_\mathrm{motor}$, occurs shortly after the transfer, when the left-hand robot halts the downward motion of the box after its release by the right-hand robot.

The corresponding plots from the dense evaluation carried out using updated nominal parameters $\bar{\bm{\theta}}'$ are included in Appendix \ref{app:dense_eval}.

\subsection{Results Sparse Safety Evaluation}
First, a sparse evaluation using the original world parameter distribution $p(\bm\theta)$ is carried out. In this, the safety score corresponding to the critical $\mathrm{FOS}_{\mathrm{motor}}$ event exceeds the safety tolerance $S > \epsilon$, indicating that the proposed action sequence may not be safe to execute. 

To demonstrate the effect of uncertainty reduction, a second dense evaluation is performed using the updated nominal values $\bm\theta'$, which reveals two new critical events corresponding to the two safety factors. When the safety scores of these events are computed using the refined, narrower parameter distribution, both satisfy $S < \epsilon$, indicating that, based on this updated knowledge, the proposed action sequence can be executed with guaranteed safely in practice.

The safety scores evaluated using both the nominal and updated parameter distributions at the critical state-transitions are presented in Table \ref{tab:safety_scores}. 
\begin{table}[b]
    \caption{Safety scores at the critical state-transitions}
    \begin{center}
    \begin{tabular}{lll}\toprule
        & Nominal: $p(\theta)$ & Updated: $p'(\theta)$\\
        \midrule
        $\mathrm{FOS}^{-1}_{\mathrm{contact}}$-event   & 0.733 & 0.319\\
        $\mathrm{FOS}^{-1}_{\mathrm{motor}}$-event & 0.781$^{\mathrm{a}}$ &  0.367 \\
        \bottomrule
        \multicolumn{3}{l}{$^{\mathrm{a}}$This score exceeds the safety tolerance $\epsilon = 0.75$.}
    \end{tabular}
    \label{tab:safety_scores}
    \end{center}
\end{table}

\begin{figure}
    \centering
    \includegraphics[width=1\linewidth]{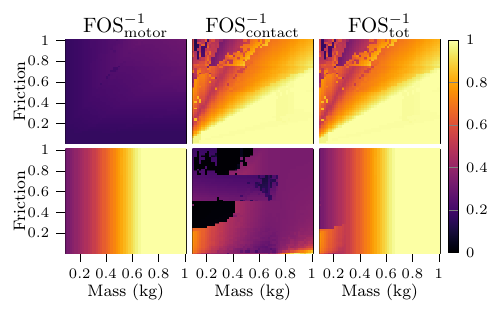}
    \caption{Factors of safety when evaluated using the original tentative parameter probability distribution $p(\bm \theta)$, with the $\mathrm{FOS}^{-1}_{\mathrm{contact}}$-event on the top and the $\mathrm{FOS}^{-1}_{\mathrm{motor}}$-event at the bottom.}
    \label{fig:nominal_FOS}
\end{figure}

\begin{figure}[h!]
    \centering
    \includegraphics[width=1\linewidth]{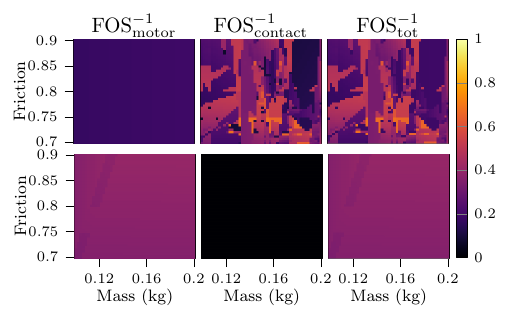}
    \caption{Factors of safety evaluated using the updated parameter distribution $p'(\theta)$. Top row: $\mathrm{FOS}^{-1}_{\mathrm{contact}}$, bottom row: $\mathrm{FOS}^{-1}_{\mathrm{motor}}$.
}
    \label{fig:updated_FOS}
\end{figure}

The monitored safety factors underlying these evaluations can be found in Fig. \ref{fig:nominal_FOS} for the nominal world parameter distribution, and in Fig. \ref{fig:updated_FOS} for the updated distribution.
\begin{figure}[tb]
    \centering
    \includegraphics[width=1\linewidth]{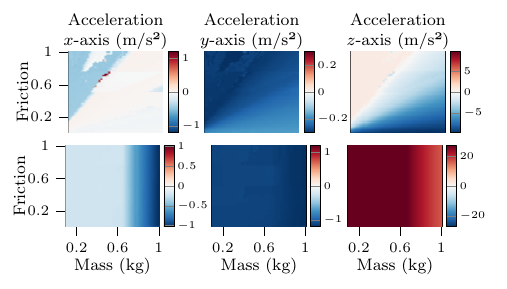}
    \caption{Accelerations in the world coordinate directions over the possible parameter distribution for the original $p(\bm \theta)$ at the two critical state-transitions, with the $\mathrm{FOS}^{-1}_{\mathrm{contact}}$-event on the top and the $\mathrm{FOS}^{-1}_{\mathrm{motor}}$-event at the bottom. The direction of gravity is along the negative $z$-axis.}
    \label{fig:accelerations}
\end{figure}

Fig. \ref{fig:accelerations} presents the acceleration of the box in world coordinate directions at these state-transitions, over the nominal parameter distribution $p(\bm \theta)$. While the accelerations are not used for the safety evaluation, they are included to illustrate how the dynamics in the system depends on the parameter values. By comparing these plots to the corresponding ones showing the inverse factors of safety, one can also observe a correlation, showing that the safety factors can be used as indications of possibly unwanted effects. By comparing the plots corresponding to the $\mathrm{FOS}^{-1}_{\mathrm{contact}}$ event in of Figs. \ref{fig:nominal_FOS} and \ref{fig:accelerations}, one can observe that a high inverted factor of safety correlates with significant acceleration in the direction of gravity. It is also apparent that the inverted safety factor obtains higher values ``before" the accelerations start to appear, since it indicates how near the system is to possible failure. Similar patterns can be observed when comparing plots showing the $\mathrm{FOS}^{-1}_{\mathrm{motor}}$ event, where the reduced acceleration against gravity for larger box masses is preceded by a high factor of safety, indicating that the actuators are operating near their limits for these world parameter values.

Fig \ref{fig:updated_FOS}, on the other hand, shows that the factors of safety are consistently small over the entire probability span after the probing action has refined it. This is reflected in the low safety scores in Table \ref{tab:safety_scores}. 

\section{Computational Feasibility Analysis}
For the investigated method to be practically useful, it must be computationally efficient enough to examine enough combinations of world parameters. The method consists of two main steps. First, dense evaluation is performed by simulated rollout from the start to the end of a control episode. Assume that we allocate a safety filter time budget equal to the duration of the control episode. Let $\tau$ denote how many times faster the simulator is than real-time in single-threaded execution. With $K$ threads, we can perform $N_d = K/\tau$ dense evaluations. If we choose to do sparse evaluations at a fraction $\alpha$ of all possible state-transitions from a dense rollout, then there is room for $N_s = K/(\alpha \tau)$ sparse evaluations with unique sets of parameter combinations. With a real-time factor of $\tau = 10$, $K = 100$ parallel simulation threads, and sparse evaluation on a fraction $\alpha = 1/1000$ of the state-transitions, it is possible to evaluate $N_d = 10^3$ nominal and $N_s = 10^6$ tentative combinations of world parameters with dense and sparse evaluation, respectively. In practice, the numbers will be lower as both steps must be carried out sequentially (divide by two) and some overhead is associated with instantiating the simulations.
As comparison, the numerical experiments of this paper were carried out with a consumer-grade desktop computer equipped with an Intel Core i7-8700K CPU and we measured a realtime factor of 1.8 without any particular efforts to optimize the code. 





\section{Discussion}

In this paper, we limited our numerical experiments to a controlled example where we only let two parameters be subject to uncertainty. Furthermore, we only identified and monitored two critical state-transitions of the entire planned action sequence. This choice was made because we wanted to limit the amount of results for clarity's sake, but in a real application one would likely need to monitor more critical state-transitions to give a more robust safety evaluation of the complete trajectory.

Additionally, we only executed one dense evaluation for each parameter probability distribution. In a real scenario, doing this could lead to a less trustworthy safety evaluation, since it makes the outcome overly reliant on that the nominal parameter values are somewhat close to the true parameter values. If, in a scenario like the one we have examined, the box had properties which were significantly different from the nominal values, the true states evolving from the control inputs might significantly deviate from the ones we analyzed. This could make the dense evaluation results less relevant, since they would not reflect the true states very well. To mitigate this issue, one could increase the number of dense evaluations, to cover a wider span of the parameter probability distribution. The exact number of dense evaluations carried out would have to be a trade-off between computational load and accuracy.

When comparing our safety evaluation method with previous works on safety filtering, one obvious difference it that our method does not have a built-in way to map a proposed unsafe action to a safe one \cite{annurev:/content/journals/10.1146/annurev-control-071723-102940, Data-driven}. This makes a separate safe policy a necessity for implementation in real life. Here further research is needed on how to formulate such a policy. 

The method relies on the assumption that we have access to a well-calibrated simulator of high fidelity. This is common to have for robots used in safety-critical missions, for example in space, mines or in nuclear facilities. They are indispensable for training operators for teleoperation and virtual commissioning of system changes prior to its real commissioning. Calibration of simulators from experimental data is challenging but there are recent tools that give automated support and can reach high precision \cite{Marklund2025}. When available, these simulators can then be used also for safety filtering without much additional cost. 

An advantage our method has over many previous works on safety filters is that it is relatively trivial to implement on new task and environments as long as one has a simulated clone of the environment. The method also allows for large freedom in what one would classify as unsafe outcomes of an action, since the FOS-functions can be tuned according to need. The method is also explainable since it is purely based on physics, and not data-driven like many other less computationally heavy safety-filtering methods are.

\section{Conclusion}
We have investigated the use of a high-fidelity simulator to evaluate whether a control policy can be safely deployed under environment uncertainty. The sparse evaluation strategy reduces the computational intensity and makes it possible to calculate how the safety factor depends on the parameter distribution of the external environment. The method we present does not provide absolute guarantees for whether a control policy is safe to deploy or not, this is the subject of further research, but it offers a new way to investigate safety and how one can act with intention to increase it.

\appendix
\subsection{Reward Definition} \label{app:reward_def}
We adopt the rewards ($r$) from the original ACT paper \cite{zhao2023learningfinegrainedbimanualmanipulation} to track the modes of the manipulation task. They are as follows:
\setlist[enumerate,1]{start=0}
\begin{enumerate}[label=\arabic*:]\setcounter{enumii}{-1}
    \item box on table, no contact with either robot 
    \item box on table, touched by gripper of the left robot 
    \item box detach from table, held by left robot gripper 
    \item box in contact with both robot grippers, free from table 
    \item box held by right robot gripper, free from table 
\end{enumerate}
\subsection{Dense Evaluation Results} \label{app:dense_eval}
\begin{figure}
\vspace{2mm}
    \centering
    \includegraphics[width=1\linewidth]{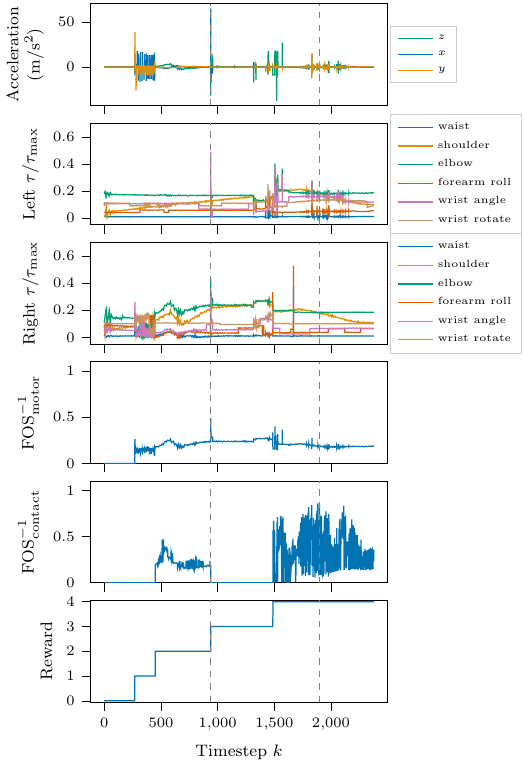}
    \caption{Results from the dense evaluation with the updated nominal parameters, corresponding to $p'(\bm \theta)$. The two critical state transitions on which the sparse evaluation is carried out are marked by gray dashed lines.}
    \label{fig:results_dense_eval_updated}
\end{figure}

Here, the monitored factors of safety from the dense evaluations, together with additional measurements included for interpretability, are provided. The results corresponding to the tentative nominal parameters and to the updated ones are provided in Figs. \ref{fig:results_dense_eval} and \ref{fig:results_dense_eval_updated}, respectively. The headings labeled $\tau/\tau_{\max}$ denote the fraction of the available torque used by the respective actuators of the left and right robots at that given time, while $a$ is the acceleration of the box given in the world coordinate directions. The reward is defined as in Appendix \ref{app:reward_def}.



\end{document}